\documentclass[conference]{IEEEtran}
\IEEEoverridecommandlockouts
\usepackage{cite}
\usepackage{amsmath,amssymb,amsfonts}
\usepackage{algorithmic}
\usepackage{graphicx}
\usepackage{textcomp}
\usepackage{xcolor}
\usepackage{lipsum}
\def\BibTeX{{\rm B\kern-.05em{\sc i\kern-.025em b}\kern-.08em
    T\kern-.1667em\lower.7ex\hbox{E}\kern-.125emX}}

\usepackage{graphicx} 
\newtheorem{remark}{Remark}
\begin{document}

\title{Convolutional Normalizing Flows for\\
Deep Gaussian Processes\thanks{This research is supported by A*STAR under its RIE$2020$ Advanced Manufacturing and Engineering (AME) Programmatic Funds (Award A$20$H$6$b$0151$).}
}
\makeatletter
\newcommand{\linebreakand}{%
  \end{@IEEEauthorhalign}
  \hfill\mbox{}\par
  \mbox{}\hfill\begin{@IEEEauthorhalign}
}
\makeatother

\author{\IEEEauthorblockN{Haibin Yu}
\IEEEauthorblockA{\textit{Tencent Data Platform} \\
Shenzhen, China \\
haibin@u.nus.edu}
\and
\IEEEauthorblockN{Dapeng Liu}
\IEEEauthorblockA{\textit{Tencent Data Platform} \\
Shenzhen China \\
rocliu@tencent.com}
\linebreakand
\IEEEauthorblockN{Yizhou Chen}
\IEEEauthorblockA{\textit{National University of Singapore} \\
Republic of Singapore \\
ychen041@comp.nus.edu.sg}
\and
\IEEEauthorblockN{Bryan Kian Hsiang Low}
\IEEEauthorblockA{\textit{National University of Singapore} \\
Republic of Singapore \\
lowkh@comp.nus.edu.sg}
\and
\IEEEauthorblockN{Patrick Jaillet}
\IEEEauthorblockA{\textit{MIT} \\
Cambridge, MA, USA \\
jaillet@mit.edu}
}

\maketitle

\begin{abstract}

\emph{Deep Gaussian processes} (DGPs), a hierarchical composition of GP models, have successfully boosted the expressive power of their single-layer counterpart. However, it is impossible to perform exact inference in DGPs, which has motivated the recent development of variational inference-based methods. Unfortunately, either these methods yield a biased posterior belief or it is difficult to evaluate their convergence. This paper introduces a new approach for specifying flexible, arbitrarily complex, and scalable approximate posterior distributions. The posterior distribution is constructed through a \emph{normalizing flow} (NF) which transforms a simple initial probability into a more complex one through a sequence of invertible transformations. Moreover, a novel \emph{convolutional normalizing flow} (CNF) is developed to improve the time efficiency and capture dependency between layers. Empirical evaluation shows that CNF DGP outperforms the state-of-the-art approximation methods for DGPs.

\end{abstract}


\section{Introduction}

\emph{Gaussian process} (GP) models \cite{Rasmussen06,NghiaAAAI19,Ruofei18,chen2013parallel,low2014parallel,MinhAAAI17,HoangICML19,NghiaICML16,HoangICML16,LowECML14a,teng20,LowAAAI14,HaibinAPP} have been widely applied in the machine learning community as they are capable of providing closed-form predictions in the form of a Gaussian distribution and formal measures of predictive uncertainty. Some examples include safety-critical applications \cite{reeb2018learning}, computer vision \cite{mark2017conv}, Bayesian optimization \cite{srinivas2009gaussian,Erik17,NghiaAAAI18,ling16,kharkovskii2020private,dmitrii20a,dai2019bayesian,zhang2017information,zhang2020bayesian,PhongAAAI21,dai20,dai20b}, active learning \cite{zimmer2018safe,LowAAMAS13,NghiaICML14,LowAAMAS08,LowICAPS09,LowAAMAS11,LowAAMAS12,LowAAMAS14,YehongAAAI16,LowTASE15,LowUAI12,chen2013gaussian,PhongAAAI21b}, among others. However, the expressiveness of GP models are limited by their kernel functions which are challenging to design and often require expert knowledge for various complex learning tasks. To this end, in recent years, we have witnessed a successful hierarchical composition of GP models into a multi-layer \emph{deep} GP (DGP) model \cite{damianou2013deep,hensman2014nested,bui2016deep,cutajar2017random,salimbeni2017doubly,havasi2018inference,yu2019ipvi}, which has boosted the expressive power significantly. Unfortunately, unlike the single-layer counterpart, DGPs do not provide tractable inference, which has motivated a series of approximate inference methods. In particular, most previous works focus on \emph{variational inference} (VI) \cite{damianou2013deep,hensman2014nested,dai2015variational,salimbeni2017doubly} by imposing Gaussian posterior assumptions. However, it has been pointed out by the work of \cite{havasi2018inference} that the posterior belief demonstrates non-Gaussian patterns, hence potentially compromising the performance of the VI methods due to biased posterior estimation. To address this, the work of \cite{havasi2018inference} proposed to utilize \emph{Markov chain Monte Carlo} (MCMC) sampling method to draw unbiased samples from the posterior belief. However, it is computational costly for generating samples in both training and prediction due to its sequential sampling procedure, let alone the difficulty in evaluating the convergence. 

To remedy the assumptions above, the work of \cite{yu2019ipvi} proposed the \emph{implicit posterior variational inference} (IPVI) framework for DGP inference which can recover the unbiased posterior distribution efficiently. The method is inspired by the generative adversarial networks \cite{goodfellow2014generative} which casts the DGP inference into a two-player game with the aim of searching for a Nash equilibrium. However, it is also mentioned in the paper that the Nash equilibrium is not guaranteed to be obtained. 

This paper proposes a novel framework which utilizes the notion of \emph{normalizing flows}  \cite{tabak2010density,tabak2013family,rezende2015variational} (NFs) which model the complex posterior distribution directly through a sequence of invertible neural networks. The benefits of choosing NFs are their expressiveness in  modeling complex real-world data distributions and the use of NFs has demonstrated success in supervised, semi-supervised, and unsupervised learning. 

\section{Background and Related work} \label{sec: background}
    \noindent
    \textbf{Gaussian Process (GP).} A GP defines a distribution over functions $f: \mathbb{R}^{D} \rightarrow \mathbb{R}$, for which any finite marginals follows a Gaussian distribution \cite{Rasmussen06}. A GP is fully specified by its mean function which is often assumed to be zero and covariance (kernel) function $k: \mathbb{R}^{D} \times \mathbb{R}^{D} \rightarrow \mathbb{R}$. Suppose that a set of $N$ inputs $\mathbf{X} \triangleq \{\mathbf{x}_n\}_{n=1}^{N}$ and their corresponding noisy outputs $\mathbf{y} \triangleq \{y_n\}_{n=1}^{N}$ are available where $y_n\triangleq f(\mathbf{x}_n)+\varepsilon$ (corrupted by an i.i.d.~Gaussian noise $\varepsilon \sim \mathcal{N}(0, \sigma_n^2)$). Then, the set of latent outputs $\mathbf{f} \triangleq \{f(\mathbf{x}_n)\}_{n=1}^{N}$ follow a Gaussian distribution $p(\mathbf{f}) = \mathcal{N}(\mathbf{0}, \mathbf{K}_{\mathbf{X}\mathbf{X}})$ where $\mathbf{K}_{\mathbf{X}\mathbf{X}}$ denotes a covariance matrix with components $k(\mathbf{x}_n, \mathbf{x}_n')$ for $n, n^\prime = 1, \dots, N$. A widely used covariance function $k(\mathbf{x}_n, \mathbf{x}_n')$ is the \emph{squared exponential} (SE) kernel with \emph{automatic relevance determination} (ARD) $k_{\theta}(\mathbf{x}, \mathbf{x}^\prime) \triangleq \sigma_f^2 \exp (-0.5 \sum_{d=1}^D (x_d - x^\prime_d)^2 / l_d^2)$ where $l_d$ is the lengthscale for the $d$-th input dimension, $\sigma_f^2$ is the kernel variance, and  $\theta \triangleq (\{l_d\}_{d=1}^D, \sigma_f)$ are the kernel hyperparameters.

    It follows that the marginal likelihood $p(\mathbf{y}) = \mathcal{N}(\mathbf{y}|\mathbf{0}, \mathbf{K}_{\mathbf{X}\mathbf{X}}+\sigma_n^2 \mathbf{I})$. The GP posterior belief for the latent outputs $\mathbf{f}^{\star} \triangleq \{f(\mathbf{x}^{\star})\}_{\mathbf{x}^{\star} \in \mathbf{X}^{\star}}$ can be computed analytically for any set $\mathbf{X}_{\star}$ of test inputs:
    \begin{equation}
        p(\mathbf{f}^{\star}|\mathbf{y}) = \int p(\mathbf{f}^{\star}|\mathbf{f}) p(\mathbf{f}|\mathbf{y})\ \mathrm{d}\mathbf{f}
        \label{eq: fgp infer}
    \end{equation}  
    which can be written as 
    \begin{equation*}
    \begin{array}{cc}
        p(\mathbf{f}^{\star}|\mathbf{y}) = \mathcal{N}(\boldsymbol{\mu}^{\star}, \boldsymbol{\Sigma}^{\star})
    \end{array}
    \end{equation*} 
    where $\boldsymbol{\mu}^{\star} \triangleq \mathbf{k}_{\mathbf{x}^\star\mathbf{X}}(\mathbf{K}_{\mathbf{X}\mathbf{X}}+ \sigma_n^2 \mathbf{I})^{-1}\mathbf{y}$ and $\boldsymbol{\Sigma}^{\star}\triangleq k_{\mathbf{x}^\star\mathbf{x}^\star} - \mathbf{k}_{\mathbf{x}^\star\mathbf{X}}(\mathbf{K}_{\mathbf{X}\mathbf{X}}+ \sigma_n^2 \mathbf{I})^{-1}\mathbf{k}_{\mathbf{X}\mathbf{x}^\star}$.
    
    Unfortunately, the inference procedure above incurs $\mathcal{O}(N^3)$ time, hence scaling poorly to massive datasets. To improve its scalability, the \emph{sparse GP} (SGP) models spanned by the unifying view of~\cite{candela05} exploit a set $\mathbf{u} \triangleq \{u_m \triangleq f(\mathbf{z}_m)\}_{m=1}^{M}$ of inducing output variables for some small set $\mathbf{Z} \triangleq \{\mathbf{z}_m\}_{m=1}^{M}$ of inducing inputs (i.e., $M \ll N$). Then,
    \begin{equation}
    p(\mathbf{y}, \mathbf{f}, \mathbf{u}) = p(\mathbf{y}|\mathbf{f})\ p(\mathbf{f}|\mathbf{u})\ p(\mathbf{u})
    \label{crap}
    \end{equation}
    such that $p(\mathbf{f}|\mathbf{u}) = \mathcal{N}(\mathbf{f}|\mathbf{K}_{\mathbf{X}\mathbf{Z}}\mathbf{K}_{\mathbf{Z}\mathbf{Z}}^{-1}\mathbf{u}, \mathbf{K}_{\mathbf{X}\mathbf{X}} - \mathbf{K}_{\mathbf{X}\mathbf{Z}}\mathbf{K}_{\mathbf{Z}\mathbf{Z}}^{-1}\mathbf{K}_{\mathbf{Z}\mathbf{X}})$
    where, with a slight abuse of notation, $\mathbf{u}$ is treated as a column vector here, $\mathbf{K}_{\mathbf{X}\mathbf{Z}}\triangleq \mathbf{K}^{\top}_{\mathbf{Z}\mathbf{X}}$, and $\mathbf{K}_{\mathbf{Z}\mathbf{Z}}$ and $\mathbf{K}_{\mathbf{Z}\mathbf{X}}$ denote covariance matrices with components $k(\mathbf{z}_m, \mathbf{z}_{m'})$ for $m,m'=1,\ldots,M$ and $k(\mathbf{z}_m, \mathbf{x}_{n})$ for $m =1,\ldots,M$ and $n=1,\ldots,N$, respectively. The SGP predictive belief can also be computed in closed form by marginalizing out $\mathbf{u}$: $p(\mathbf{f}^\star|\mathbf{y}) = \int p(\mathbf{f}^\star|\mathbf{u})\ p(\mathbf{u}|\mathbf{y})\ \mathrm{d}\mathbf{u}$.

    The work of \cite{Titsias09} proposed a principled \emph{variational inference} (VI) framework that  approximates the joint posterior belief $p(\mathbf{f}, \mathbf{u}|\mathbf{y})$ with a variational posterior $q(\mathbf{f}, \mathbf{u}) \triangleq p(\mathbf{f}|\mathbf{u})\ q(\mathbf{u})$ by minimizing the \emph{Kullback-Leibler} (KL) divergence between them, 
    which is equivalent to maximizing a lower bound of the log-marginal likelihood (i.e., also known as the \emph{evidence lower bound} (ELBO)):
        \begin{equation*}
            \mathrm{ELBO} \triangleq \mathbb{E}_{q(\mathbf{f})}[\log p(\mathbf{y}|\mathbf{f})] - \mathrm{KL}[q(\mathbf{u})\Vert p(\mathbf{u})]
        \end{equation*}
    where $q(\mathbf{f}) \triangleq \int p(\mathbf{f}|\mathbf{u})\ q(\mathbf{u})\ \mathrm{d}\mathbf{u}$.\vspace{1mm} 
        

\noindent
    \textbf{Deep Gaussian Process (DGP).} A DGP model composes multiple layers of GP models. Consider a DGP model with a depth of $L$ such that each DGP layer is associated with a set $\mathbf{F}_{\ell-1}$ of inputs and a set $\mathbf{F}_{\ell}$ of outputs for $\ell = 1, \dots, L$ and $\mathbf{F}_0 \triangleq \mathbf{X}$. Let $\boldsymbol{\mathcal{F}}\triangleq\{\mathbf{F}_{\ell}\}_{\ell=1}^L$, and the inducing inputs and corresponding inducing output variables for DGP layers $\ell = 1, \dots, L$ be denoted by the respective sets $\boldsymbol{\mathcal{Z}}\triangleq\{\mathbf{Z}_{\ell}\}_{\ell=1}^L$ and $\boldsymbol{\mathcal{U}}\triangleq\{\mathbf{U}_{\ell}\}_{\ell=1}^L$. Similar to the joint probability distribution of the SGP model in~\eqref{crap}, the joint probability distribution of DGP can be written as 
    \begin{equation*}
        p(\mathbf{y}, \boldsymbol{\mathcal{F}}, \boldsymbol{\mathcal{U}})
        =  \underbrace{p(\mathbf{y}|\mathbf{F}_L)}_\text{data likelihood}\;
        \underbrace{\left[\prod_{\ell=1}^{L}p(\mathbf{F}_{\ell}|\mathbf{U}_{\ell})\right]p(\boldsymbol{\mathcal{U}})}_\text{DGP prior}.
    \end{equation*}
Similarly, the variational posterior is assumed to be $q(\boldsymbol{\mathcal{F}}, \boldsymbol{\mathcal{U}}) \triangleq \left[\prod_{\ell=1}^{L}p(\mathbf{F}_{\ell}|\mathbf{U}_{\ell})\right]q(\boldsymbol{\mathcal{U}})$, thus resulting in the following ELBO for the DGP model:
        \begin{equation}
        \vspace{-2mm}
            \mathrm{ELBO} \triangleq \int q(\mathbf{F}_L) \log p(\mathbf{y}|\mathbf{F}_L)\ \mathrm{d}\mathbf{F}_L  - \mathrm{KL}[q(\boldsymbol{\mathcal{U}})\Vert p(\boldsymbol{\mathcal{U}})] 
            \label{eq: elboDGP}
        \end{equation}
        where  \vspace{-0mm}
        $$q(\mathbf{F}_L) \triangleq \int \prod_{\ell=1}^{L}p(\mathbf{F}_{\ell}|\mathbf{U}_{\ell},\mathbf{F}_{\ell-1})\ q(\boldsymbol{\mathcal{U}})\ \mathrm{d}\mathbf{F}_1\dots \mathrm{d}\mathbf{F}_{L-1}\mathrm{d}\boldsymbol{\mathcal{U}}\ .$$ 
        Note that $q(\mathbf{F}_L)$ is computed using the reparameterization trick proposed in the work of \cite{kingma2013auto} and Monte Carlo sampling method proposed in the work of \cite{salimbeni2017doubly}.


    	Previous VI frameworks for DGP models \cite{damianou2013deep,hensman2014nested,dai2015variational,salimbeni2017doubly} have imposed the restrictive Gaussian mean-field assumption on $q(\boldsymbol{\mathcal{U}})$. Unfortunately, it has been pointed by the work of~\cite{havasi2018inference} that the true posterior distribution of $q(\boldsymbol{\mathcal{U}})$ usually exhibits a high correlation across the DGP layers and is non-Gaussian, hence potentially compromising the performance of such VI-based DGP models. To further remove these assumptions, \cite{yu2019ipvi} proposed the IPVI framework for DGP that can capture the dependency between layers and ideally recover the unbiased posterior distribution. To achieve this, the method casts the DGP inference problem as a two-player game and search for a Nash equilibrium using \emph{best response dynamics} (BRD).\footnote{This procedure is sometimes called “better-response dynamics” (http://timroughgarden.org/f13/l/l16.pdf).}  However, the works of \cite{goodfellow2016nips,salimans2016improved} have pointed out the critical issue of convergence while searching for Nash equilibrium. In fact, \cite{yu2019ipvi} also mentioned that there is no guarantee for BRD to converge to a Nash equilibrium, hence giving no assurance of recovering the true posterior distribution. 
    	
    	The goal of DGP inference lies in three aspects: recovery of the true posterior, convergence analysis, and time efficiency. To achieve this, this paper proposes a novel framework based on the idea of \emph{convolutional normalizing flows} (CNF) to model the posterior distribution directly and efficiently, as detailed in Section~\ref{sec: nf for DGP}.

    
    \section{Convolutional Normalizing Flow for DGPs} \label{sec: nf for DGP}
\noindent
    \textbf{Normalizing Flows.} By examining the ELBO in Eq.~\ref{eq: elboDGP} in detail, we can find out that the maximum of the ELBO is achieved when $\mathrm{KL}[q(\boldsymbol{\mathcal{U}})\Vert p(\boldsymbol{\mathcal{U}}|\mathbf{y})] = 0$. Our goal is to develop an ideal family of variational distributions $q(\boldsymbol{\mathcal{U}})$ that is  flexible enough to represent the true posterior distribution. We introduce the notion of \emph{normalizing flow}\cite{tabak2010density,tabak2013family} (NF) here. An NF describes the transformation of a simple distribution into a complex distribution by repeatedly applying a sequence of invertible mappings. Following the \emph{change of variable} rule, the NF framework can be described as follows: Given a random variable $\mathbf{z} \in \mathbb{R}^D$ with distribution $\pi(\mathbf{z})$, there exists an invertible and smooth mapping $f: \mathbb{R}^D \rightarrow \mathbb{R}^D$. Then, the resulting random variable $\mathbf{x} = f(\mathbf{z})$ has the distribution
    \begin{equation*}
        p(\mathbf{x}) = \pi(\mathbf{z})\left|\mathrm{det}\left(\frac{\mathrm{d}\mathbf{z}}{\mathrm{d}\mathbf{x}}\right)\right| = \pi(\mathbf{z})\ |\mathrm{det}[f^{\prime}(\mathbf{z})]|^{-1}
    \end{equation*}
    where $|\mathrm{det}[f^{\prime}(\mathbf{z})]|^{-1}$ is the absolute value of the determinant of the Jacobian of $f$ evaluated at $\mathbf{z}$.\vspace{1mm} 
\noindent
    \begin{remark}
    It has been proven that certain NFs are universal approximators \cite{huang2018neural,jaini2019tails}. In other words, it means that with a careful design of the mapping $f$, we can transform a simple distribution into any complex distribution.\footnote{We refer the readers to \cite{villani2003topics,bogachev2005triangular,medvedev2008certain,huang2018neural,jaini2019tails} for a detailed discussion of the proof.}\vspace{1mm} 
    \end{remark}
%
 %
    Recall that maximizing the ELBO is equivalent to minimizing the KL divergence, specifically, the KL divergence between our variational posterior $q(\boldsymbol{\mathcal{U}})$ and the true posterior $p(\boldsymbol{\mathcal{U}}|\mathbf{y})$. Following the \emph{Bayes' theorem}, the posterior distribution can be written as
    \begin{equation}
        p(\boldsymbol{\mathcal{U}}|\mathbf{y}) = \dfrac{p(\mathbf{y}|\mathbf{F}_{L}, \boldsymbol{\mathcal{U}})\ p(\boldsymbol{\mathcal{U}})}{p(\mathbf{y}|\mathbf{F}_{L})}\ . 
    \end{equation}
    Similarly, $\mathbf{F}_{L}$ is a Monte Carlo sample from the predictive distribution of the layer outputs:  $$\mathbf{F}_{L} \sim \int \prod_{\ell=1}^{L-1}p(\mathbf{F}_{\ell}|\mathbf{U}_{\ell}, \mathbf{F}_{\ell - 1})\  \mathrm{d}\mathbf{F}_{\ell}\ .$$ 
    Inspired by the idea of NF, we propose to construct the variational posterior $q(\boldsymbol{\mathcal{U}})$ through a sequence of invertible and smooth mappings: $\mathcal{G} (\boldsymbol{\mathcal{V}}) = \boldsymbol{\mathcal{U}}$ where $\boldsymbol{\mathcal{V}}$ is a new random variable with distribution $\pi(\cdot)$. Then, according to the \emph{change of variable} rule,
    \begin{equation}
        q(\boldsymbol{\mathcal{U}}) = \pi(\boldsymbol{\mathcal{V}}) \times \left|\mathrm{det}\left(\dfrac{\mathrm{d}\mathcal{G}}{\mathrm{d}\boldsymbol{\mathcal{V}}}\right)\right|^{-1}.
    \label{eq: change of variable}
    \end{equation}
    Using \eqref{eq: change of variable}, the KL divergence can be re-written as
    \begin{equation}
    \hspace{-1mm}
    \begin{array}{l}
        \displaystyle \mathrm{KL}(q(\boldsymbol{\mathcal{U}})\Vert p(\boldsymbol{\mathcal{U}}|\mathbf{y})) = \mathrm{KL}\left(q(\boldsymbol{\mathcal{U}})\Big\Vert\dfrac{p(\mathbf{y}|\mathbf{F}_{\ell}, \boldsymbol{\mathcal{U}})\ p(\boldsymbol{\mathcal{U}})}{p(\mathbf{y}|\mathbf{F}_{\ell})}\right) \vspace{1mm}
        \\
        \displaystyle= \int q(\boldsymbol{\mathcal{U}}) \left[\log \dfrac{q(\boldsymbol{\mathcal{U}})}{p(\mathbf{y}|\mathbf{F}_{\ell}, \boldsymbol{\mathcal{U}})\ p(\boldsymbol{\mathcal{U}})}\right] \mathrm{d}\boldsymbol{\mathcal{U}} + \log p(\mathbf{y}|\mathbf{F}_{\ell}) \vspace{1mm}
        \\
        \displaystyle = \mathbb{E}_{q(\boldsymbol{\mathcal{U}})} \left[\log q(\boldsymbol{\mathcal{U}})- \log p(\mathbf{y}|\mathbf{F}_{\ell}, \boldsymbol{\mathcal{U}})-p(\boldsymbol{\mathcal{U}})\right] + \log p(\mathbf{y}|\mathbf{F}_{\ell}) \vspace{1mm}
        \\
        \displaystyle= \mathbb{E}_{\pi(\boldsymbol{\mathcal{V}})}\bigg[\log \pi(\boldsymbol{\mathcal{V}}) - \log\left|\mathrm{det}\left(\dfrac{\mathrm{d} \mathcal{G}}{\mathrm{d} \boldsymbol{\mathcal{V}}}\right)\right| -\log p(\mathbf{y}|\mathbf{F}_{\ell}, \mathcal{G}(\boldsymbol{\mathcal{V}})) 
        \\
        \displaystyle \quad- \log p(\mathcal{G}(\boldsymbol{\mathcal{V}}))\bigg] + \mathrm{const}\ .
    \end{array}
    \label{eq: kl divergence}
    \end{equation}
    \begin{remark}
    Note that the IPVI DGP framework in \cite{yu2019ipvi} implicitly represents the variational posterior  with samples. Hence, to compute the log-density ratio in the KL divergence, it resorts to the use of a \emph{discriminator} to output a function value to represent it. However, in our NF framework, the log-density for the variational posterior $\log q(\boldsymbol{\mathcal{U}}) = \log \pi(\boldsymbol{\mathcal{V}}) - \log\left|\mathrm{det}\left(\mathrm{d} \mathcal{G}/\mathrm{d} \boldsymbol{\mathcal{V}}\right)\right|$ results in an analytical solution.\vspace{1mm} 
    \end{remark}



\noindent
\textbf{Convolutional Normalizing Flows.} We will now discuss how the architecture of the normalizing flow is designed for DGP. A naive design is to consider a layer-wise normalizing flow which is illustrated in Fig.~\ref{fig: nf naive}. However, such a naive design suffers from the following critical issues:
            \begin{figure}[ht]
                \centering 
                \includegraphics[width=80mm]{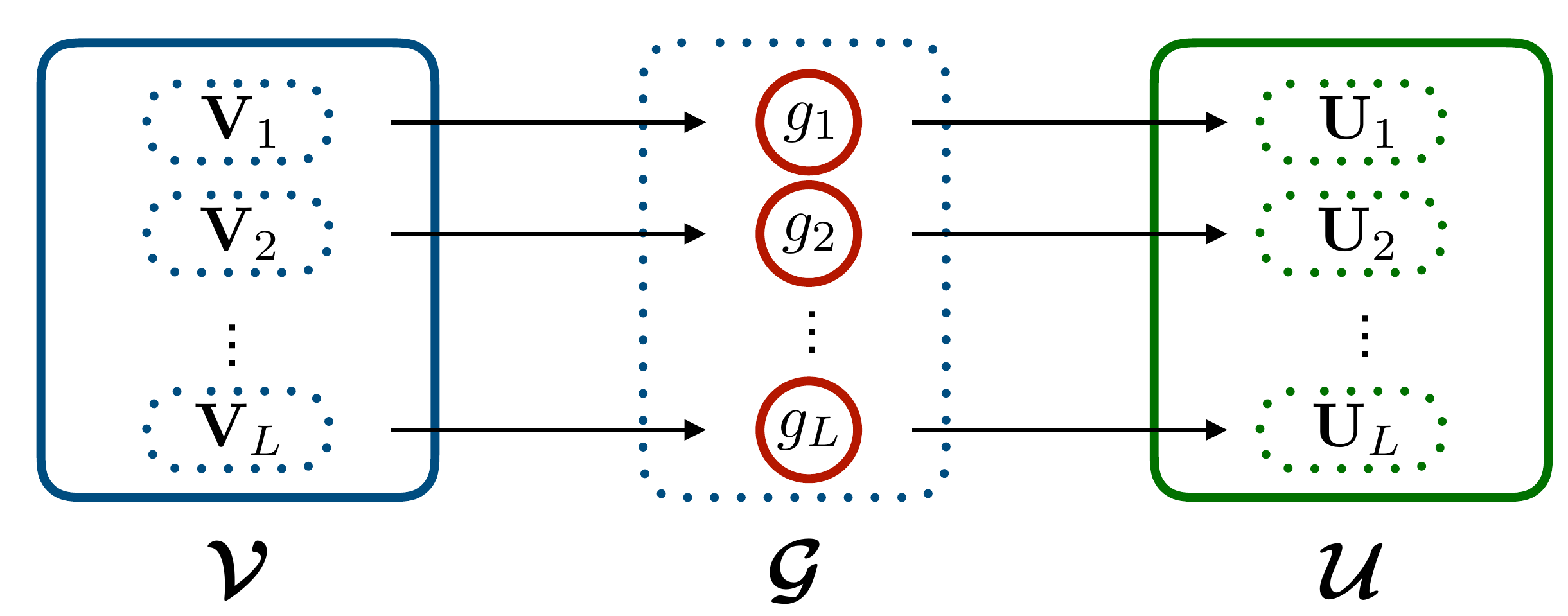}
                \caption{A naive design of normalizing flow for DGP. The normalizing flow $\boldsymbol{\mathcal{G}}$ is separated into $L$ individual flows. }
                \label{fig: nf naive}
            \end{figure}
\begin{itemize}
\item Fig.~\ref{fig: nf naive} shows that to recover the posterior samples of $M$ different inducing variables 
$\mathbf{U}_{\ell}\triangleq\{\mathbf{u}_{\ell 1}, \dots, \mathbf{u}_{\ell M}\}$ where $\mathbf{u}_{\ell 1}, \dots, \mathbf{u}_{\ell M} \in \mathbb{R}^{d_{\ell}}$, it is natural to design the normalizing flow $g_{\ell}: \mathbb{R}^{Md_{\ell}} \rightarrow \mathbb{R}^{Md_{\ell}}$. Therefore, a large number of parameters is needed, which will increase the risk of overfitting; 
\item Another critical issue is the computational complexity. In general, computing the log-Jacobian determinant incurs $\mathcal{O}(M^3 d_{\ell}^3)$ time, hence resulting in the difficulty in optimization; 
\item Such a design fails to capture the dependency of the inducing output variables $\mathbf{U}_{\ell}$ among different layers. As pointed out by \cite{yu2019ipvi}, the posterior distribution of $p(\boldsymbol{\mathcal{U}}|\mathbf{y})$ is highly correlated among layers; 
\item Such a design fails to adequately capture the dependency of the inducing output variables $\mathbf{U}_{\ell}$ on its corresponding inducing inputs $\mathbf{Z}_{\ell}$, hence restricting its capability to model the output posterior $\boldsymbol{\mathcal{U}}$ accurately. 
\end{itemize}
            To resolve the above issues, we propose a novel normalizing flow architecture with convolution for DGP models, as shown in Fig.~\ref{fig: nf novel}. Instead of treating $\boldsymbol{\mathcal{V}} \triangleq \{\mathbf{V}_1, \mathbf{V}_2, \dots, \mathbf{V}_L \}$ separately, we decide to stack $\{\mathbf{V}_1, \mathbf{V}_2, \dots, \mathbf{V}_L \}$ to be a three-dimensional tensor denoted as
            \begin{equation}
                \boldsymbol{\mathcal{V}} \in \mathbb{R}^{M \times 1 \times \sum_{\ell=1}^{L}d_{\ell}}
            \end{equation}
            and design our normalizing flow $\boldsymbol{\mathcal{G}}: \mathbb{R}^{M \times 1 \times \sum_{\ell=1}^{L}d_{\ell}} \rightarrow \mathbb{R}^{M \times 1 \times \sum_{\ell=1}^{L}d_{\ell}}$ accordingly, as shown in Fig.~\ref{fig: nf novel}. 
            \begin{figure}[ht]
                \centering 
                \includegraphics[width=65mm]{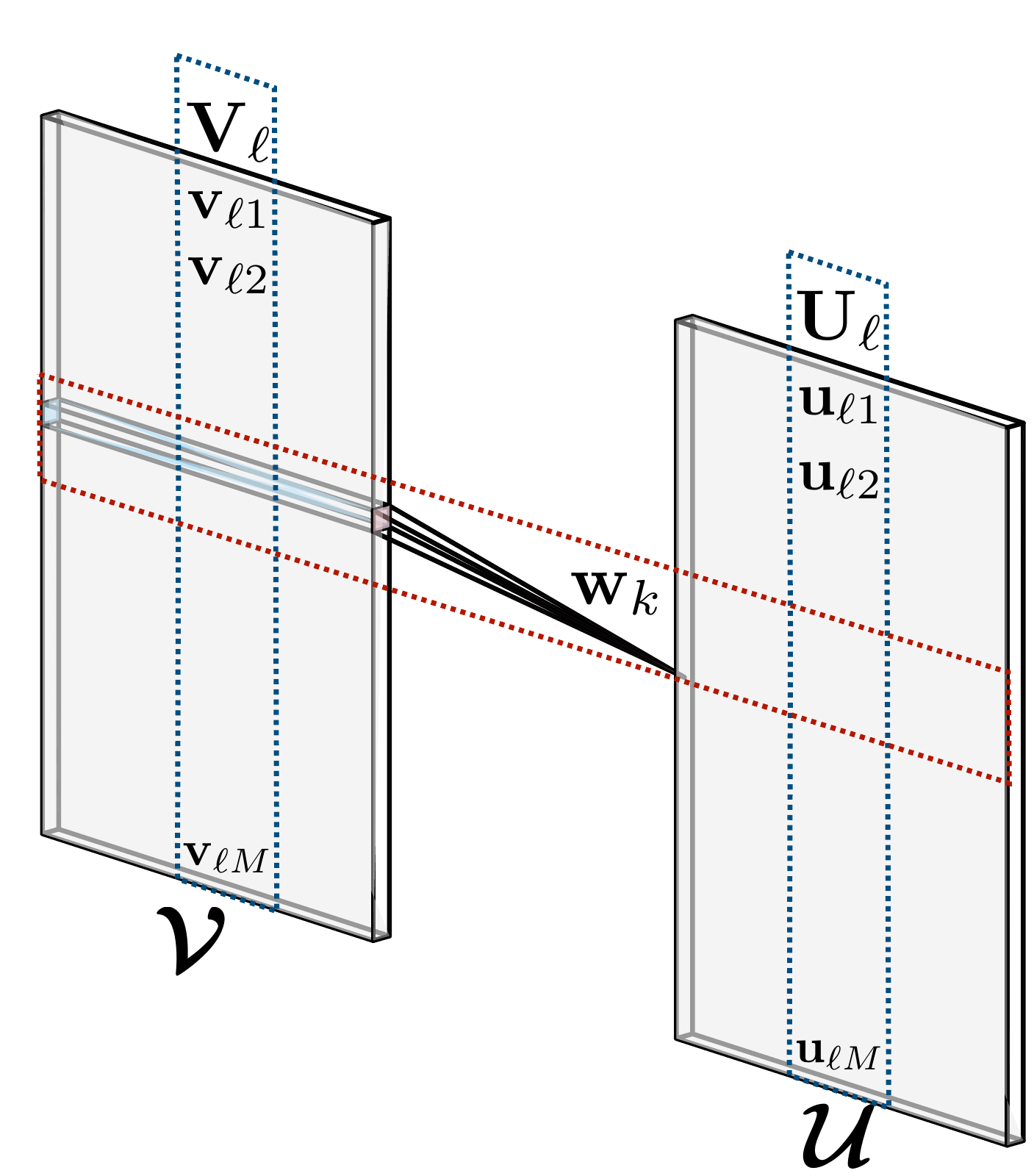}
                \caption{The normalizing flow with convolution for DGP. The kernel tensor $\mathbf{W}$ can be decomposed into a set of tensors $\{\mathbf{w}_1, \mathbf{w}_2, \dots, \mathbf{w}_K\}$ where $\mathbf{w}_1, \dots, \mathbf{w}_K \in \mathbb{R}^{1\times 1\times \sum_{\ell=1}^{L}d_{\ell}}$ and
                $K \triangleq \sum_{\ell=1}^{L}d_{\ell}$. 
                The red box indicates the convolution with $\mathbf{w}_k$.}
                \label{fig: nf novel}
            \end{figure}
            To this end, we propose to convolve $\boldsymbol{\mathcal{V}}$ with a kernel tensor $\mathbf{W}$ where $\mathbf{W} \in \mathbb{R}^{1 \times 1 \times \sum_{\ell=1}^{L}d_{\ell} \times \sum_{\ell=1}^{L}d_{\ell}}$, as shown in Fig.~\ref{fig: nf novel}. In this manner, the normalizing flow $\mathcal{G}$ is fully characterized by the kernel tensor $\mathbf{W}$ with a significantly smaller number of parameters. Hence, the log-Jacobian determinant can be easily written as
            \begin{equation}
                \log \left|\mathrm{det} \left(\dfrac{\mathrm{d}\ \mathcal{G}}{\mathrm{d}\ \boldsymbol{\mathcal{V}}}\right)\right| = M \times 1 \times \log|\mathrm{det}(\mathbf{W})|\ .\vspace{1mm}
            \end{equation}

            \begin{remark}
             Note that compared with the naive design in Fig.~\ref{fig: nf naive}, computing the log-Jacobian determinant only incurs $\mathcal{O}((\sum_{\ell=1}^{L}d_{\ell})^{3})$ time, which reduces the time complexity by a factor of $\mathcal{O}(M^3)$. Moreover, compared with the naive design, it is easier to compute the determinant of $\mathbf{W}$. Another advantage is that the kernel tensor $\mathbf{W}$ naturally captures the dependency of inducing variables $\{ \mathbf{U}_1, \mathbf{U}_2, \dots, \mathbf{U}_L \}$ between layers.\vspace{1mm} 
            \end{remark}
            Furthermore, to capture the dependency of the inducing output variables $\mathbf{U}_{\ell}$ on its corresponding inducing inputs $\mathbf{Z}_{\ell}$, we manually construct the base distribution $\pi(\boldsymbol{\mathcal{V}})$ to depend on the inducing inputs $\boldsymbol{\mathcal{Z}}$. Specifically, for each layer $\ell$, the base distribution can be written as
            \begin{equation}
                \mathbf{V}_{\ell} \sim \mathcal{N}(\boldsymbol{\mu}_{\ell}, \boldsymbol{\sigma}_{\ell}), \quad [\boldsymbol{\mu}_{\ell}, \boldsymbol{\sigma}_{\ell}] = \varphi_{\ell}(\mathbf{Z}_{\ell})
            \end{equation}
            where $\varphi_{\ell}$ is a neural network. We represent $\varphi_{\ell}$ using a two-layer neural network with the dimension of the hidden layer being $256$ and leaky ReLU activation in the hidden units. Note that it utilizes a separate set of parameters for each different layer $\ell$. We observe from our experiments that our normalizing flow for DGP with convolutions improves the performance considerably, which will be shown in Section~\ref{sec: experiments}.

\section{Experiments and Discussions} \label{sec: experiments}
    \subsection{Regression} \label{subsec: regression}

        \noindent\textbf{UCI Benchmark Regression.} Our experiments are first conducted on $7$ UCI benchmark regression datasets. We have performed a random $0.9/0.1$ train/test split.
    
    \begin{figure}[ht]
    \hspace{2mm}
    \begin{tabular}{lllll}
        \hspace{-6mm} \includegraphics[height=30mm]{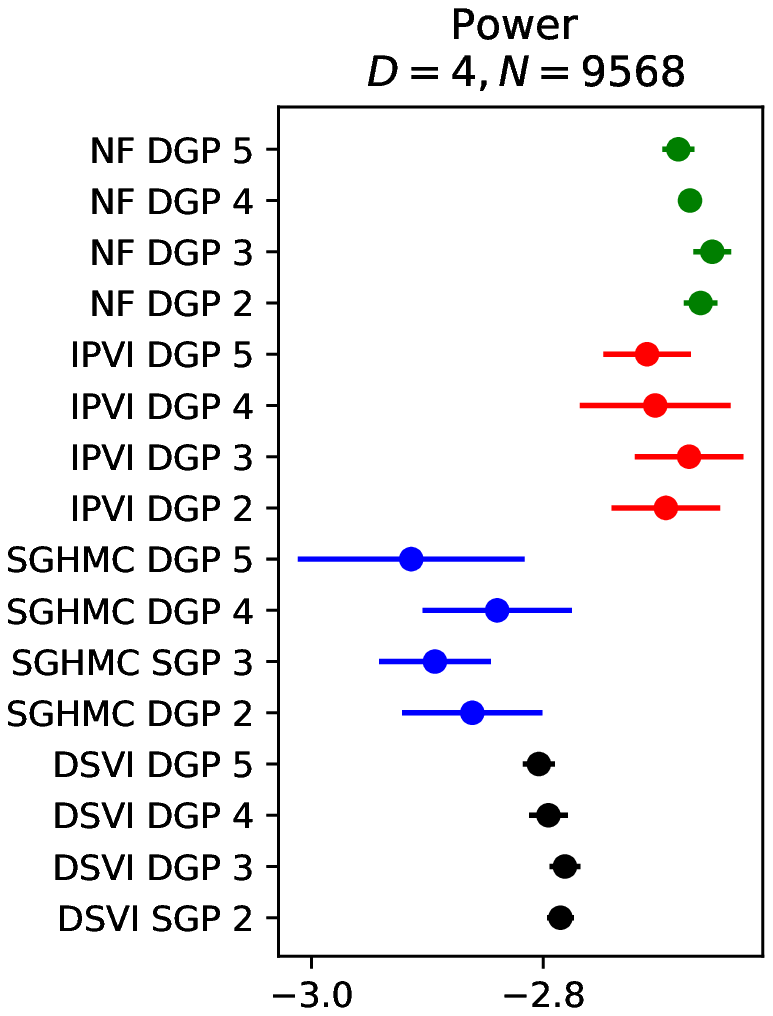}
        & \hspace{-6mm} \includegraphics[height=30mm]{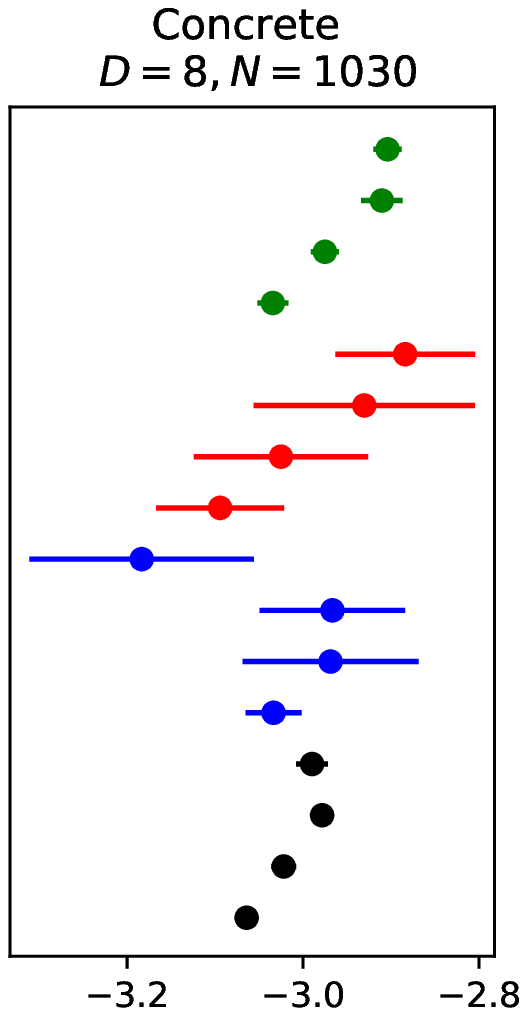}
        & \hspace{-6mm} \includegraphics[height=30mm]{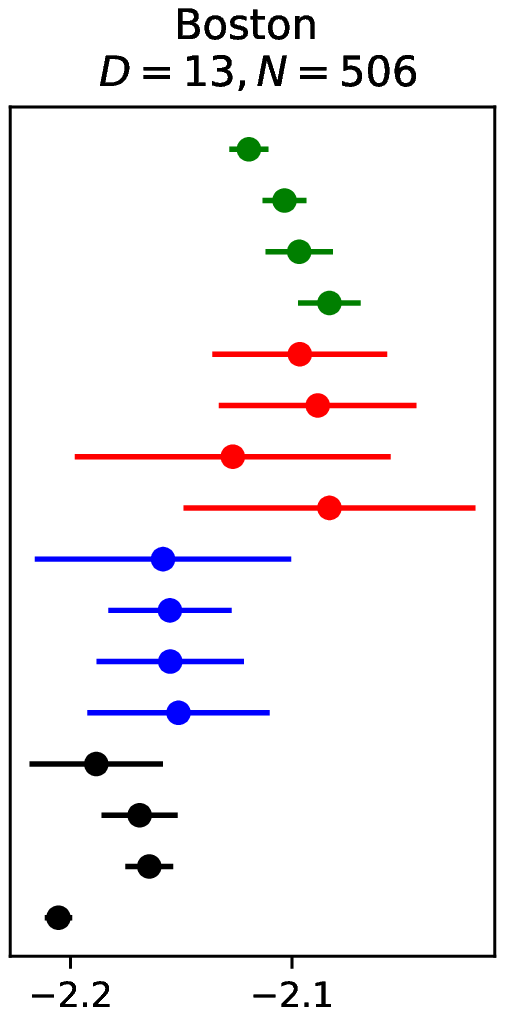}
        & \hspace{-6mm} \includegraphics[height=30mm]{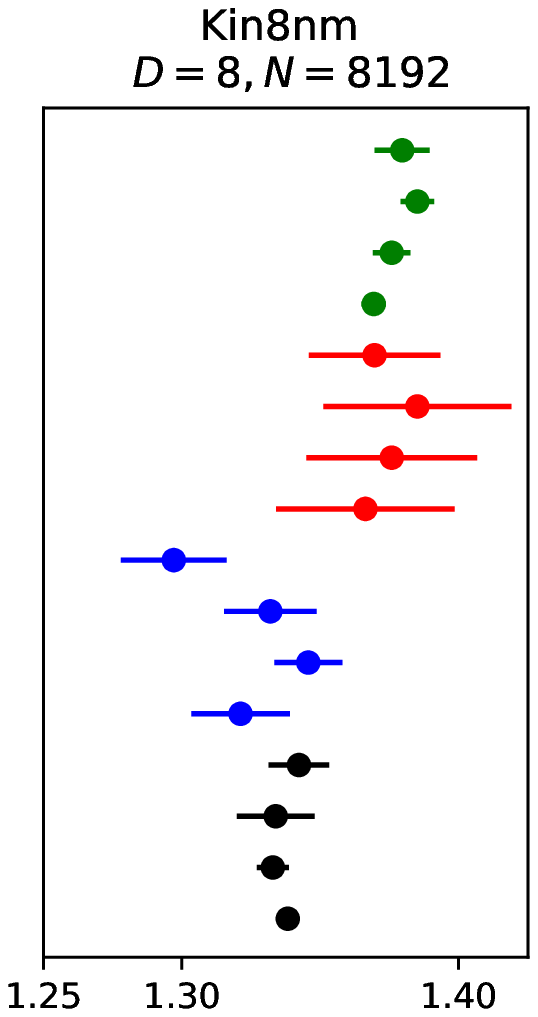}
        & \hspace{-6mm} \includegraphics[height=30mm]{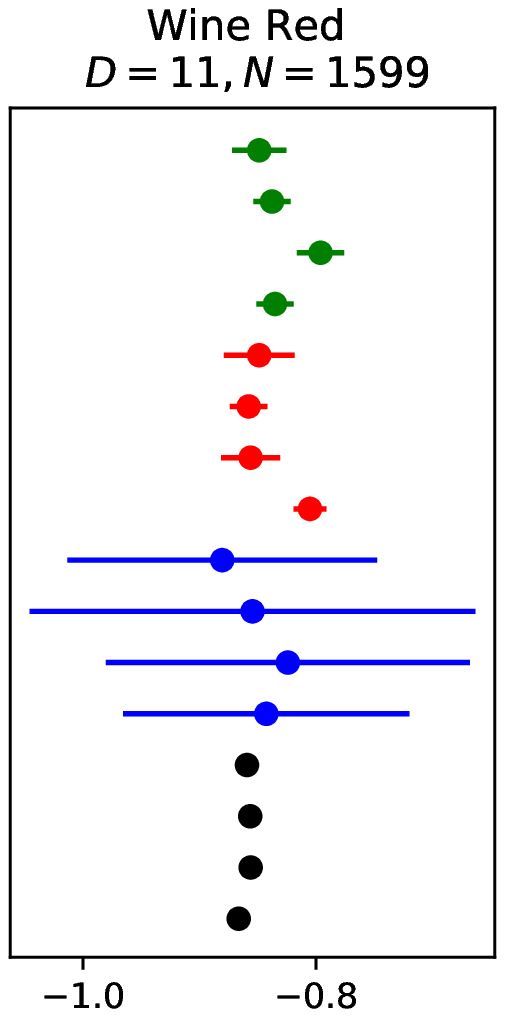}
        \vspace{-1.5mm}
        \\
         \hspace{-6mm} \includegraphics[height=30mm]{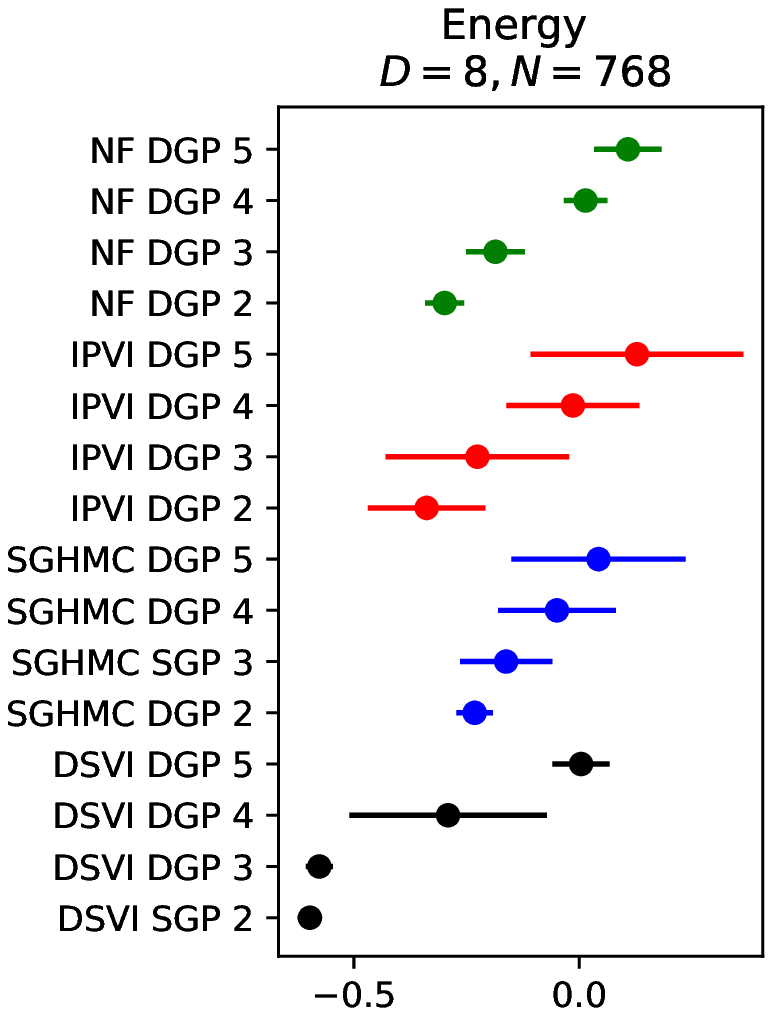}
        & \hspace{-6mm} \includegraphics[height=30mm]{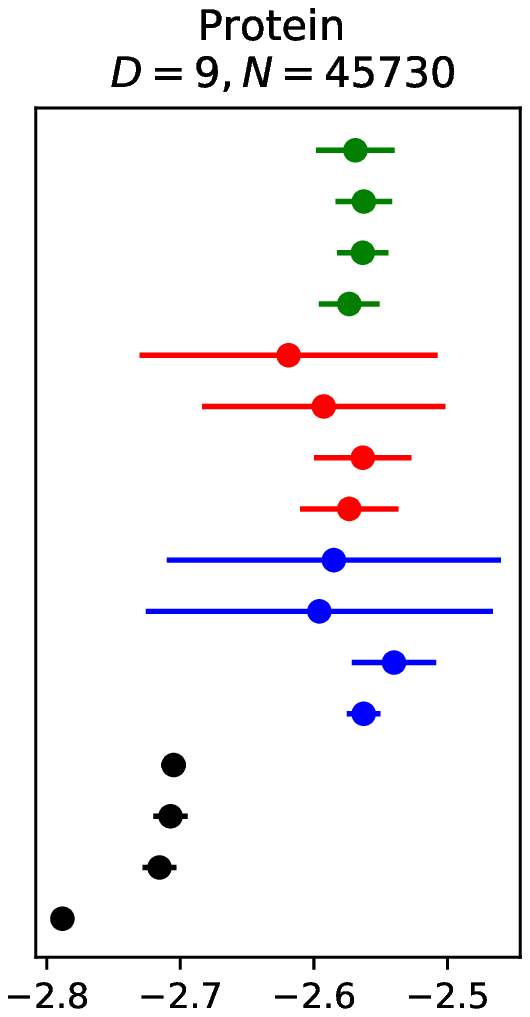}
        & \hspace{-6mm} \includegraphics[height=30mm]{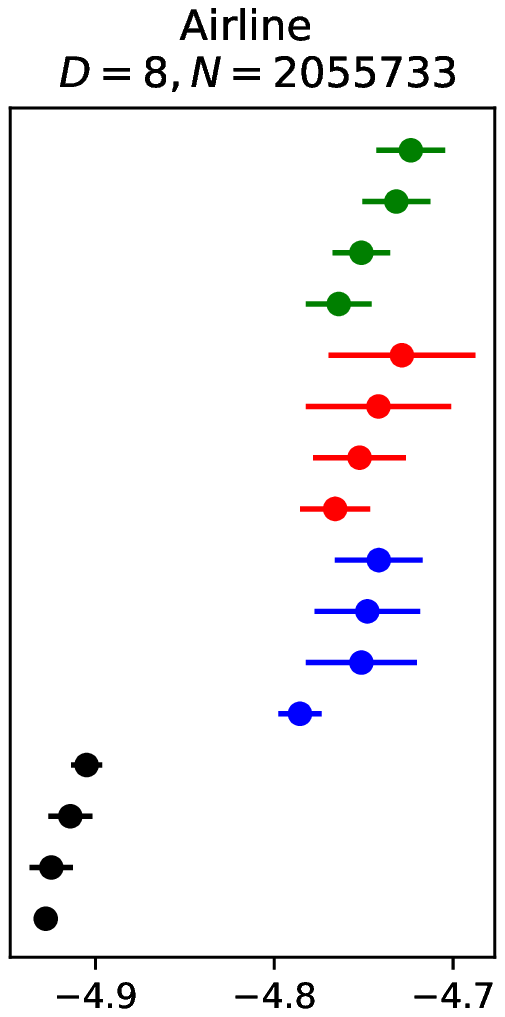}
        & \hspace{-6mm} \includegraphics[height=30mm]{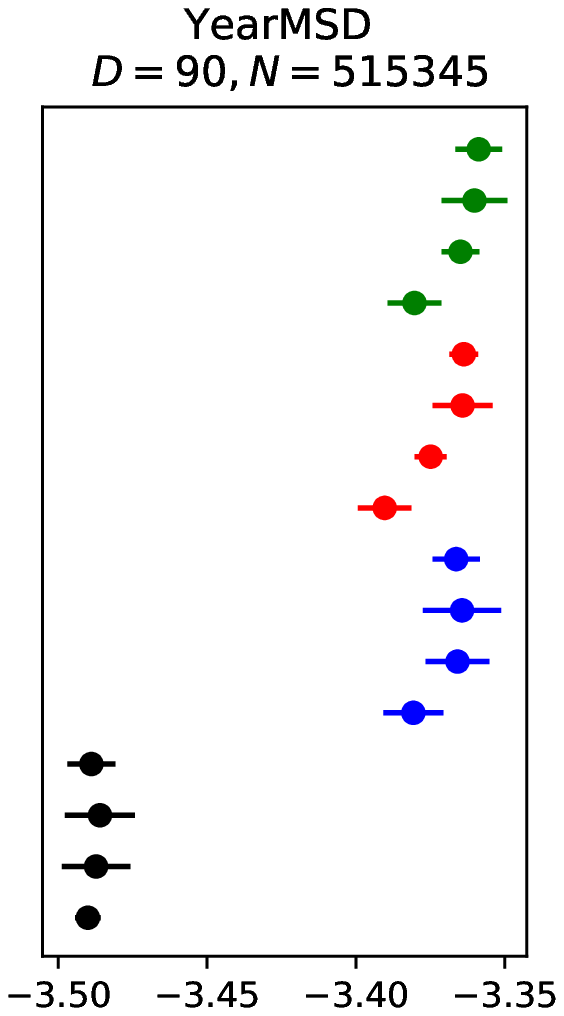}
        & \hspace{-6mm} \includegraphics[height=30mm]{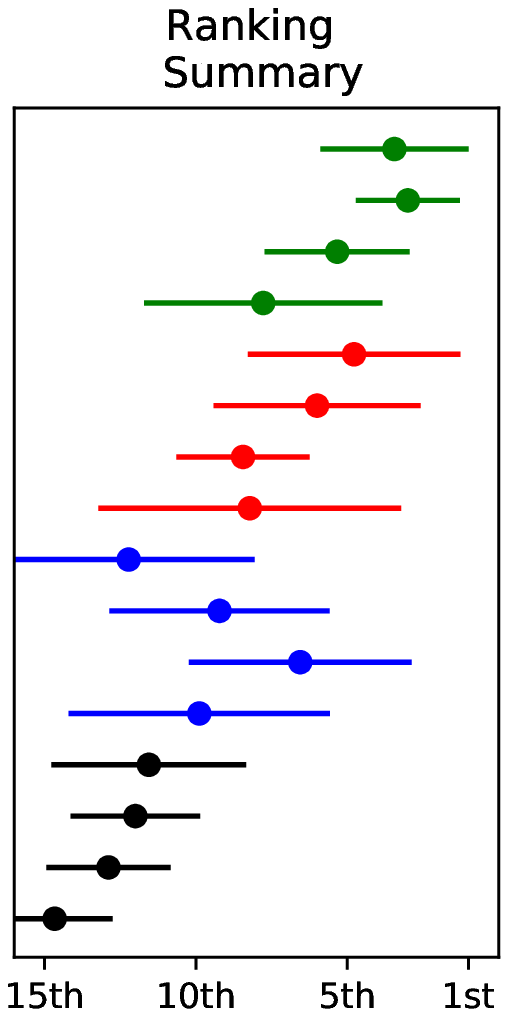}
        \vspace{-1.5mm}
        \end{tabular}
    \caption{Mean test log-likelihood and standard deviation achieved by our NF framework (green), IPVI (red), SGHMC (blue), and DSVI (black) for DGPs for UCI benchmark and large-scale regression datasets. Higher test log-likelihood (i.e., to the right) is better.}
    \label{fig: uci}\vspace{-1mm} 
    \end{figure}
        
    \noindent\textbf{Large-Scale Regression.} We then evaluate the performance of NF on two real-world large-scale regression datasets: (a) YearMSD dataset with a large input dimension $D=90$ and data size $N\approx 500000$, and (b) Airline dataset with input dimension $D=8$ and a large data size $N\approx2$~million.
    For YearMSD dataset, we use the first $463715$ examples as training data and the last $51630$ examples as test data.
    For Airline dataset, we set the last $100000$ examples as test data. 

    In the above regression tasks, the performance metric is the \emph{mean test log-likelihood} (MLL). Fig.~\ref{fig: uci} shows the results of the mean test log-likelihood and standard deviation over $10$ runs. It can be observed that NF generally outperforms other frameworks and the ranking summary shows that our NF framework for a $4$-layer DGP model (NF DGP 4) performs the best on average across all regression tasks. For large-scale regression tasks, the performance of NF consistently increases with a greater depth.\vspace{1mm} 


    
    \noindent\textbf{Evaluation of ELBO.} To further demonstrate the expressiveness of the NF framework, we have computed the estimate of training ELBO for NF DGP and IPVI DGP models on Boston dataset. Table~\ref{tab: elbo} shows the mean ELBOs of NF and IPVI over $10$ runs for the Boston dataset. NF generally achieves higher ELBOs, which agrees with results of the test MLL in Fig.~\ref{fig: uci}.
\begin{table}[ht]
    \centering
    \caption{Mean ELBOs for Boston dataset.}
    \label{tab: elbo}
    \begin{tabular}{|c|cc|}
    \hline
    Model & \multicolumn{1}{c}{NF} & IPVI \\ \hline
    DGP 2          & {-836.48}                        & {-846.65}    \\
    DGP 3          & {-814.13}                        & {-846.45}    \\
    DGP 4          & {-762.54}                        & {-776.93}    \\
    DGP 5          & {-734.23}                        & {-758.42}    \\ \hline
    \end{tabular}
\end{table}

    \begin{remark}
        As can be observed from Fig.~\ref{fig: uci}, NF DGP is very robust across different runs (as can be seen from the smaller standard deviations) compared with SGHMC and IPVI which represent the posterior distribution with samples. Moreover, Table~\ref{tab: elbo} clearly demonstrates that NF works better in recovering the true posterior distribution, as compared with IPVI. The inferior performance for IPVI is due to the convergence issue mentioned previously.\vspace{1mm} 
    \end{remark}

    \noindent\textbf{Time efficiency.} Table~\ref{tab: time} shows the time efficiency of NF DGP, as compared with IPVI DGP and SGHMC DGP. 

        \begin{table}[ht]
        \hspace{-3mm}
        \caption{Time incurred by a $5$-layer DGP model for Airline dataset.}
        \centering
        \begin{tabular}{cccc}
            \hline
            & NF DGP  & IPVI DGP & SGHMC DGP \\
            \hline
            Average training time & 0.56 sec  & 0.42 sec  & 3.67 sec \\
            \hline 
            Average generation time & 0.32 sec & 0.31 sec & 156.2 sec
            \\
            \hline
        \end{tabular}   
        \label{tab: time}       
        \end{table}
    \begin{remark}
        As can be observed from Table~\ref{tab: time}, the average training time and generation time for NF DGP are slightly longer than IPVI DGP. Since NF DGP can achieve more superior predictive performance (Fig~\ref{fig: uci}) as it does not suffer from convergence issues, this is an acceptable trade-off. 
    \end{remark}
    \subsection{Classification} \label{subsec: classification}
    We evaluate the performance of NF in three classification tasks using the real-world MNIST, fashion-MNIST, and CIFAR-$10$ datasets. Both MNIST and fashion-MNIST datasets are gray-scale images of $28\times28$ pixels. The CIFAR-$10$ dataset consists of colored images of $32\times32$ pixels. We utilize a $4$-layer DGP model with $100$ inducing inputs per layer and a robust-max multiclass likelihood~\cite{hernandez2011robust}.\vspace{1mm}
    
    \noindent\textbf{Convolutional Skip-layer Connection (CSC).} For the image datasets, the data distribution has local correlation between pixels. It would be better if the skip-layer connection can incorporate such information as a base for invariant mapping. We change the skip-layer connection from a fully connected one to a convolutional one; $\mathbf{W}$ is a convolution kernel with height and width of $3\times 3$. Note that in CSC, $\mathbf{W}$ is trainable. The results in Table~\ref{tab: nf classifications conv} show that the CSC boosts the DGP performance in real-world image datasets and performs best when integrated with the NF framework.

        \begin{table}[ht]
        \centering
        \hspace{0mm}\caption{Mean test accuracy (\%) achieved by NF, IPVI, SGHMC, and DSVI for $3$ classification datasets with convolutions.}
        \begin{tabular}{l|ll|ll|ll}
        \hline
        Dataset     & \multicolumn{2}{c|}{MNIST}                        & \multicolumn{2}{c|}{Fashion-MNIST}                & \multicolumn{2}{c}{CIFAR-10}                      \\ \hline
                    & SGP            & DGP 4                     & SGP            & DGP 4         & SGP            & DGP 4                     \\ \hline
        DSVI        & 97.32 & 99.16               & 86.98          & 91.57                   & 47.15          & 75.05                   \\
        SGHMC       & 96.41          & 98.15                    & 85.84          & 88.14                   & 47.32          & 70.78                  \\
        IPVI & 97.02          & 99.32 & \textbf{87.29} & 91.78  & 48.07 & 76.11 \\
        \textbf{NF} & \textbf{97.37} & \textbf{99.43}  & 87.14 & \textbf{92.03}  & \textbf{48.13} & \textbf{76.81}  \\
        \hline
        \end{tabular}   
        \label{tab: nf classifications conv}
        \end{table}
        
    \section{Conclusion} \label{sec: conclusion} 
        This paper proposes a novel NF framework for DGP inference which is targeted at resolving issues of the biased mean-field Gaussian approximation as well as convergence in IPVI. We also propose a novel \emph{convolutional NF} (CNF) architecture for DGPs to better capture dependency between inducing variables and speed up training. Empirical evaluation shows that CNF performs better than other state-of-the-art approximation methods for DGPs in regression and classification tasks. For future work, we plan to extend the CNF framework for DGP to perform semi-supervised learning by incorporating the generative expressiveness of \emph{normalizing flows}.

\bibliographystyle{IEEEtran}
\bibliography{IEEEabrv, Reference}

\begin{thebibliography}{10}
\providecommand{\url}[1]{#1}
\csname url@samestyle\endcsname
\providecommand{\newblock}{\relax}
\providecommand{\bibinfo}[2]{#2}
\providecommand{\BIBentrySTDinterwordspacing}{\spaceskip=0pt\relax}
\providecommand{\BIBentryALTinterwordstretchfactor}{4}
\providecommand{\BIBentryALTinterwordspacing}{\spaceskip=\fontdimen2\font plus
\BIBentryALTinterwordstretchfactor\fontdimen3\font minus
  \fontdimen4\font\relax}
\providecommand{\BIBforeignlanguage}[2]{{%
\expandafter\ifx\csname l@#1\endcsname\relax
\typeout{** WARNING: IEEEtran.bst: No hyphenation pattern has been}%
\typeout{** loaded for the language `#1'. Using the pattern for}%
\typeout{** the default language instead.}%
\else
\language=\csname l@#1\endcsname
\fi
#2}}
\providecommand{\BIBdecl}{\relax}
\BIBdecl

\bibitem{Rasmussen06}
C.~E. Rasmussen and C.~K. Williams, \emph{{Gaussian} processes for machine
  learning}.\hskip 1em plus 0.5em minus 0.4em\relax MIT Press, 2006.

\bibitem{NghiaAAAI19}
T.~N. Hoang, Q.~M. Hoang, K.~H. Low, and J.~P. How, ``Collective online
  learning of {Gaussian} processes in massive multi-agent systems,'' in
  \emph{Proc. {AAAI}}, 2019.

\bibitem{Ruofei18}
R.~Ouyang and K.~H. Low, ``Gaussian process decentralized data fusion meets
  transfer learning in large-scale distributed cooperative perception,'' in
  \emph{Proc. AAAI}, 2018, pp. 3876--3883.

\bibitem{chen2013parallel}
J.~Chen, N.~Cao, K.~H. Low, R.~Ouyang, C.~K.-Y. Tan, and P.~Jaillet, ``Parallel
  gaussian process regression with low-rank covariance matrix approximations,''
  in \emph{Proc. {UAI}}, 2013, pp. 152--161.

\bibitem{low2014parallel}
K.~H. Low, J.~Yu, J.~Chen, and P.~Jaillet, ``Parallel gaussian process
  regression for big data: Low-rank representation meets markov
  approximation,'' in \emph{AAAI}, 2015, pp. 2821--2827.

\bibitem{MinhAAAI17}
Q.~M. Hoang, T.~N. Hoang, and K.~H. Low, ``A generalized stochastic variational
  {Bayesian} hyperparameter learning framework for sparse spectrum {Gaussian}
  process regression,'' in \emph{Proc. {AAAI}}, 2017, pp. 2007--2014.

\bibitem{HoangICML19}
Q.~M. Hoang, T.~N. Hoang, B.~K.~H. Low, and C.~Kingsford, ``Collective model
  fusion for multiple black-box experts,'' in \emph{Proc. ICML}, 2019, pp.
  2742--2750.

\bibitem{NghiaICML16}
T.~N. Hoang, Q.~M. Hoang, and B.~K.~H. Low, ``A unifying framework of anytime
  sparse {Gaussian} process regression models with stochastic variational
  inference for big data,'' in \emph{Proc. {ICML}}, 2015, pp. 569--578.

\bibitem{HoangICML16}
------, ``A distributed variational inference framework for unifying parallel
  sparse {Gaussian} process regression models,'' in \emph{Proc. ICML}, 2016,
  pp. 382--391.

\bibitem{LowECML14a}
B.~K.~H. Low, N.~Xu, J.~Chen, K.~K. Lim, and E.~B. {\"{O}zg\"{u}l},
  ``Generalized online sparse {Gaussian} processes with application to
  persistent mobile robot localization,'' in \emph{Proc. {ECML/PKDD Nectar
  Track}}, 2014, pp. 499--503.

\bibitem{teng20}
T.~Teng, J.~Chen, Y.~Zhang, and B.~K.~H. Low, ``Scalable variational {Bayesian}
  kernel selection for sparse {Gaussian} process regression,'' in \emph{Proc.
  {AAAI}}, 2020, pp. 5997--6004.

\bibitem{LowAAAI14}
N.~Xu, K.~H. Low, J.~Chen, K.~K. Lim, and E.~B. Ozgul, ``{GP-Localize}:
  Persistent mobile robot localization using online sparse {Gaussian} process
  observation model,'' in \emph{Proc. {AAAI}}, 2014, pp. 2585--2592.

\bibitem{HaibinAPP}
H.~Yu, T.~N. Hoang, B.~K.~H. Low, and P.~Jaillet, ``Stochastic variational
  inference for {Bayesian} sparse {Gaussian} process regression,'' in
  \emph{Proc. {IJCNN}}, 2019.

\bibitem{reeb2018learning}
D.~Reeb, A.~Doerr, S.~Gerwinn, and B.~Rakitsch, ``Learning {G}aussian processes
  by minimizing {P}ac-{B}ayesian generalization bounds,'' in \emph{Proc.
  NeurIPS}, 2018, pp. 3337--3347.

\bibitem{mark2017conv}
M.~van~der Wilk, C.~E. Rasmussen, and J.~Hensman, ``Convolutional {Gaussian}
  processes,'' in \emph{Proc. NeurIPS}, 2017, pp. 2849--2858.

\bibitem{srinivas2009gaussian}
N.~Srinivas, A.~Krause, S.~M. Kakade, and M.~Seeger, ``Gaussian process
  optimization in the bandit setting: No regret and experimental design,''
  \emph{Proc. ICML}, pp. 1015--1022, 2010.

\bibitem{Erik17}
E.~A. Daxberger and K.~H. Low, ``Distributed batch {Gaussian} process
  optimization,'' in \emph{Proc. {ICML}}, 2017, pp. 951--960.

\bibitem{NghiaAAAI18}
T.~N. Hoang, Q.~M. Hoang, and K.~H. Low, ``Decentralized high-dimensional
  {Bayesian} optimization with factor graphs,'' in \emph{Proc. {AAAI}}, 2018,
  pp. 3231--3238.

\bibitem{ling16}
C.~K. Ling, K.~H. Low, and P.~Jaillet, ``{Gaussian} process planning with
  {Lipschitz} continuous reward functions: Towards unifying {Bayesian}
  optimization, active learning, and beyond,'' in \emph{Proc. {AAAI}}, 2016,
  pp. 1860--1866.

\bibitem{kharkovskii2020private}
D.~Kharkovskii, Z.~Dai, and B.~K.~H. Low, ``Private outsourced {Bayesian}
  optimization,'' in \emph{Proc. {ICML}}, 2020.

\bibitem{dmitrii20a}
D.~Kharkovskii, C.~K. Ling, and B.~K.~H. Low, ``Nonmyopic {Gaussian} process
  optimization with macro-actions,'' in \emph{Proc. AISTATS}, 2020, pp.
  4593--4604.

\bibitem{dai2019bayesian}
Z.~Dai, H.~Yu, B.~K.~H. Low, and P.~Jaillet, ``Bayesian optimization meets
  {Bayesian} optimal stopping,'' in \emph{Proc. {ICML}}, 2019, pp. 1496--1506.

\bibitem{zhang2017information}
Y.~Zhang, T.~N. Hoang, B.~K.~H. Low, and M.~Kankanhalli, ``Information-based
  multi-fidelity {Bayesian} optimization,'' in \emph{Proc. {NeurIPS} Workshop
  on {Bayesian} Optimization}, 2017.

\bibitem{zhang2020bayesian}
Y.~Zhang, Z.~Dai, and B.~K.~H. Low, ``Bayesian optimization with binary
  auxiliary information,'' in \emph{Proc. {UAI}}, 2020, pp. 1222--1232.

\bibitem{PhongAAAI21}
Q.~P. Nguyen, S.~Tay, B.~K.~H. Low, and P.~Jaillet, ``Top-$k$ ranking
  {Bayesian} optimization,'' in \emph{Proc. {AAAI}}, 2021.

\bibitem{dai20}
Z.~Dai, B.~K.~H. Low, and P.~Jaillet, ``Federated {Bayesian} optimization via
  {Thompson} sampling,'' in \emph{Proc. NeurIPS}, 2020, pp. 9687--9699.

\bibitem{dai20b}
Z.~Dai, Y.~Chen, B.~K.~H. Low, P.~Jaillet, and T.-H. Ho, ``{R2-B2}: Recursive
  reasoning-based {Bayesian} optimization for no-regret learning in games,'' in
  \emph{Proc. ICML}, 2020, pp. 2291--2301.

\bibitem{zimmer2018safe}
C.~Zimmer, M.~Meister, and D.~Nguyen-Tuong, ``Safe active learning for
  time-series modeling with {G}aussian processes,'' in \emph{Proc. NeurIPS},
  2018, pp. 2730--2739.

\bibitem{LowAAMAS13}
N.~Cao, K.~H. Low, and J.~M. Dolan, ``Multi-robot informative path planning for
  active sensing of environmental phenomena: A tale of two algorithms,'' in
  \emph{Proc. AAMAS}, 2013, pp. 7--14.

\bibitem{NghiaICML14}
T.~N. Hoang, K.~H. Low, P.~Jaillet, and M.~Kankanhalli, ``Nonmyopic
  $\epsilon$-{B}ayes-optimal active learning of {Gaussian} processes,'' in
  \emph{Proc. {ICML}}, 2014, pp. 739--747.

\bibitem{LowAAMAS08}
K.~H. Low, J.~M. Dolan, and P.~Khosla, ``Adaptive multi-robot wide-area
  exploration and mapping,'' in \emph{Proc. AAMAS}, 2008, pp. 23--30.

\bibitem{LowICAPS09}
------, ``Information-theoretic approach to efficient adaptive path planning
  for mobile robotic environmental sensing,'' in \emph{Proc. ICAPS}, 2009, pp.
  233--240.

\bibitem{LowAAMAS11}
------, ``Active {Markov} information-theoretic path planning for robotic
  environmental sensing,'' in \emph{Proc. AAMAS}, 2011, pp. 753--760.

\bibitem{LowAAMAS12}
K.~H. Low, J.~Chen, J.~M. Dolan, S.~Chien, and D.~R. Thompson, ``Decentralized
  active robotic exploration and mapping for probabilistic field classification
  in environmental sensing,'' in \emph{Proc. {AAMAS}}, 2012, pp. 105--112.

\bibitem{LowAAMAS14}
R.~Ouyang, K.~H. Low, J.~Chen, and P.~Jaillet, ``Multi-robot active sensing of
  non-stationary {Gaussian} process-based environmental phenomena,'' in
  \emph{Proc. AAMAS}, 2014, pp. 573--580.

\bibitem{YehongAAAI16}
Y.~Zhang, T.~N. Hoang, K.~H. Low, and M.~Kankanhalli, ``Near-optimal active
  learning of multi-output {G}aussian processes,'' in \emph{Proc. {AAAI}},
  2016, pp. 2351--2357.

\bibitem{LowTASE15}
J.~Chen, K.~H. Low, P.~Jaillet, and Y.~Yao, ``Gaussian process decentralized
  data fusion and active sensing for spatiotemporal traffic modeling and
  prediction in mobility-on-demand systems,'' \emph{{IEEE} Transactions on
  Automation Science and Engineering}, vol.~12, no.~3, pp. 901--921, 2015.

\bibitem{LowUAI12}
J.~Chen, K.~H. Low, C.~K.-Y. Tan, A.~Oran, P.~Jaillet, J.~Dolan, and
  G.~Sukhatme, ``Decentralized data fusion and active sensing with mobile
  sensors for modeling and predicting spatiotemporal traffic phenomena,'' in
  \emph{Proc. UAI}, 2012, pp. 163--173.

\bibitem{chen2013gaussian}
J.~Chen, K.~H. Low, and C.~K.~Y. Tan, ``Gaussian process-based decentralized
  data fusion and active sensing for mobility-on-demand system,'' in
  \emph{Proc. {RSS}}, 2013.

\bibitem{PhongAAAI21b}
Q.~P. Nguyen, B.~K.~H. Low, and P.~Jaillet, ``An information-theoretic
  framework for unifying active learning problems,'' in \emph{Proc. {AAAI}},
  2021.

\bibitem{damianou2013deep}
A.~Damianou and N.~Lawrence, ``Deep {Gaussian} processes,'' in \emph{Proc.
  {AISTATS}}, 2013, pp. 207--215.

\bibitem{hensman2014nested}
J.~Hensman and N.~D. Lawrence, ``Nested variational compression in deep
  {Gaussian} processes,'' {arXiv}:1412.1370, 2014.

\bibitem{bui2016deep}
T.~Bui, D.~Hern{\'a}ndez-Lobato, J.~Hernandez-Lobato, Y.~Li, and R.~Turner,
  ``Deep {Gaussian} processes for regression using approximate expectation
  propagation,'' in \emph{Proc. {ICML}}, 2016, pp. 1472--1481.

\bibitem{cutajar2017random}
K.~Cutajar, E.~V. Bonilla, P.~Michiardi, and M.~Filippone, ``Random feature
  expansions for deep {Gaussian} processes,'' in \emph{Proc. ICML}, 2017, pp.
  884--893.

\bibitem{salimbeni2017doubly}
H.~Salimbeni and M.~Deisenroth, ``Doubly stochastic variational inference for
  deep {Gaussian} processes,'' in \emph{Proc. NeurIPS}, 2017, pp. 4588--4599.

\bibitem{havasi2018inference}
M.~Havasi, J.~M. Hern{\'a}ndez-Lobato, and J.~J. Murillo-Fuentes, ``Inference
  in deep {Gaussian} processes using stochastic gradient {Hamiltonian Monte
  Carlo},'' in \emph{Proc. NeurIPS}, 2018, pp. 7517--7527.

\bibitem{yu2019ipvi}
H.~Yu, Y.~Chen, Z.~Dai, K.~H. Low, and P.~Jaillet, ``Implicit posterior
  variational inference for deep {Gaussian} processes,'' in \emph{Proc.
  NeurIPS}, 2019, pp. 14\,475--14\,486.

\bibitem{dai2015variational}
Z.~Dai, A.~Damianou, J.~Gonz{\'a}lez, and N.~Lawrence, ``Variational
  auto-encoded deep {Gaussian} processes,'' in \emph{Proc. ICLR}, 2016.

\bibitem{goodfellow2014generative}
I.~Goodfellow, J.~Pouget-Abadie, M.~Mirza, B.~Xu, D.~Warde-Farley, S.~Ozair,
  A.~Courville, and Y.~Bengio, ``Generative adversarial nets,'' in \emph{Proc.
  NeurIPS}, 2014, pp. 2672--2680.

\bibitem{tabak2010density}
E.~G. Tabak and E.~Vanden-Eijnden, ``Density estimation by dual ascent of the
  log-likelihood,'' \emph{Communications in Mathematical Sciences}, vol.~8,
  no.~1, pp. 217--233, 2010.

\bibitem{tabak2013family}
E.~G. Tabak and C.~V. Turner, ``A family of nonparametric density estimation
  algorithms,'' \emph{Communications on Pure and Applied Mathematics}, vol.~66,
  no.~2, pp. 145--164, 2013.

\bibitem{rezende2015variational}
D.~Rezende and S.~Mohamed, ``Variational inference with normalizing flows,'' in
  \emph{Proc. ICML}, 2015, pp. 1530--1538.

\bibitem{candela05}
J.~Qui{\~n}onero-Candela and C.~E. Rasmussen, ``A unifying view of sparse
  approximate {Gaussian} process regression,'' \emph{The Journal of Machine
  Learning Research}, vol.~6, pp. 1939--1959, 2005.

\bibitem{Titsias09}
M.~Titsias, ``Variational learning of inducing variables in sparse {Gaussian}
  processes,'' in \emph{Proc. {AISTATS}}, 2009, pp. 567--574.

\bibitem{kingma2013auto}
D.~P. Kingma and M.~Welling, ``Auto-encoding variational {Bayes},'' in
  \emph{Proc. ICLR}, 2013.

\bibitem{goodfellow2016nips}
I.~Goodfellow, ``{NIPS} 2016 tutorial: Generative adversarial networks,''
  {arXiv}:1701.00160, 2016.

\bibitem{salimans2016improved}
T.~Salimans, I.~Goodfellow, W.~Zaremba, V.~Cheung, A.~Radford, and X.~Chen,
  ``Improved techniques for training gans,'' in \emph{Proc. NeurIPS}, 2016, pp.
  2234--2242.

\bibitem{huang2018neural}
C.-W. Huang, D.~Krueger, A.~Lacoste, and A.~Courville, ``Neural autoregressive
  flows,'' in \emph{Proc. ICML}, 2018, pp. 2078--2087.

\bibitem{jaini2019tails}
P.~Jaini, I.~Kobyzev, M.~Brubaker, and Y.~Yu, ``Tails of triangular flows,''
  {arXiv}:1907.04481, 2019.

\bibitem{villani2003topics}
C.~Villani, \emph{Topics in optimal transportation}.\hskip 1em plus 0.5em minus
  0.4em\relax American Mathematical Society, 2003, no.~58.

\bibitem{bogachev2005triangular}
V.~I. Bogachev, A.~V. Kolesnikov, and K.~V. Medvedev, ``Triangular
  transformations of measures,'' \emph{Matematicheskii Sbornik}, vol. 196,
  no.~3, pp. 3--30, 2005.

\bibitem{medvedev2008certain}
K.~V. Medvedev, ``Certain properties of triangular transformations of
  measures,'' \emph{Theory of Stochastic Processes}, vol.~14, no.~1, pp.
  95--99, 2008.

\bibitem{hernandez2011robust}
D.~Hern{\'a}ndez-Lobato, J.~M. Hern{\'a}ndez-Lobato, and P.~Dupont, ``Robust
  multi-class {Gaussian} process classification,'' in \emph{Proc. NeurIPS},
  2011, pp. 280--288.

\end{thebibliography}

\end{document}